\documentclass[runningheads]{llncs}

\usepackage{eccv}

\usepackage{eccvabbrv}

\usepackage{graphicx}
\usepackage{booktabs}
\usepackage{siunitx}
\DeclareSIUnit{\linepair}{lp}
\usepackage[accsupp]{axessibility}  

\usepackage[hidelinks]{hyperref}

\usepackage{orcidlink}

\usepackage{enumitem}
\usepackage{mathtools}
\usepackage{overpic}
\usepackage{amssymb}
\usepackage{xcolor}
\newcommand{\suppref}[1]{Supp. Mat. Sec.~#1}

\begin{document}

\title{Compact Low-Cost Hyperspectral Imaging\texorpdfstring{\\}{ }via Angular-to-Spectral Diversity Conversion} 

\titlerunning{Compact Low-Cost HSI via Angular-to-Spectral Diversity Conversion}


\newcommand\correspondingfootnote{
  \begingroup
  \renewcommand{\thefootnote}{\dag}
  \renewcommand{\theHfootnote}{correspondingauthor}
  \begin{NoHyper}
  \footnotetext{Corresponding author.}
  \end{NoHyper}
  \endgroup
}
\author{Kazuma Fujiwara\textsuperscript{\dag}\inst{1}\orcidlink{0009-0007-2989-5169} \and
Takuya Funatomi\inst{1,2}\orcidlink{0000-0001-5588-5932} \and
Kazuya Kitano\inst{1}\orcidlink{0000-0001-6430-6963}
\and 
Yuki~Fujimura\inst{1}\orcidlink{0000-0002-7225-8452}
\and
Yasuhiro Mukaigawa\inst{1}\orcidlink{0000-0001-8689-3724}
}
\authorrunning{K.~Fujiwara et al.}


\institute{
Nara Institute of Science and Technology, Nara, Japan
\and
Kyoto University, Kyoto, Japan\\
\email{\{fujiwara.kazuma.fj4, kitano.kazuya, fujimura.yuki, mukaigawa\}@is.naist.jp}, 
\email{funatomi.takuya.2c@kyoto-u.ac.jp}
}

\maketitle
\correspondingfootnote

\begin{abstract}
Snapshot hyperspectral imaging avoids sequential scanning, but systems that jointly achieve stable reconstruction, low cost, and compact optics remain limited.
We present a snapshot hyperspectral imaging system based on angular-to-spectral diversity conversion.
A tapered kaleidoscope creates replicated views with distinct incidence directions, and a directly attached birefringent filter converts them into view-channel-dependent spectral transmittances, yielding complementary measurements that better condition the inverse problem for more stable single-shot spectral reconstruction.
The system preserves a simple pixel-wise linear model for fast non-learning-based reconstruction and uses only off-the-shelf components without relay optics or cascaded modules.
We select the birefringent filter configuration using a condition-number-based criterion and validate the system on both synthetic and real data.
\keywords{Snapshot Hyperspectral Imaging \and Angular-to-Spectral Diversity Conversion \and Kaleidoscope \and Birefringent Filter}
\end{abstract}
\newcommand{\contribbullet}{\raisebox{0.2ex}{\Large$\bullet$}}

\section{Introduction}
\begin{figure}[tb]
    \centering
    \includegraphics[width=\textwidth]{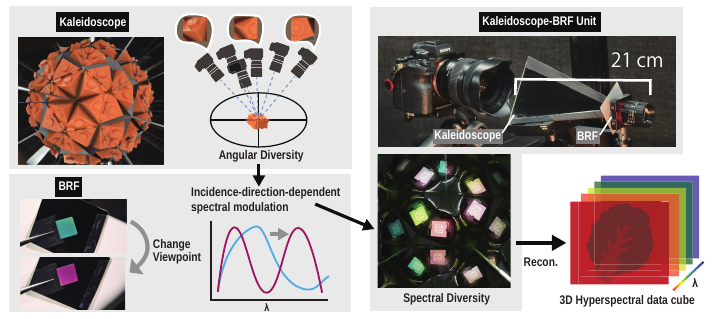}
\caption{A kaleidoscope produces angularly diverse replicated views, and an attached BRF converts them into spectral-response diversity for snapshot HSI.}
\label{fig:tether}
\end{figure}

Hyperspectral imaging (HSI) captures 3D spectral data cubes, providing much denser spectral information than the human eye or an RGB camera. HSI has been widely used in computer vision tasks such as object recognition \cite{Okajima2000_Object_Recognition}, color consistency \cite{Abrardo_1999_Color_constancy}, and anomaly detection \cite{Stein_2002_Anomaly_detection}. 
Conventionally, HSI is acquired by scanning, sequentially capturing 2D slices and incurring long capture times and complex systems.

In contrast, snapshot HSI \cite{reviewHSI_2013,ReviewsnapHSI_2023} reconstructs a spectral data cube from a single exposure, avoiding sequential scanning and simplifying acquisition.
However, snapshot HSI systems that simultaneously achieve high reconstruction performance, low cost, and compact optics remain limited, which hinders broad adoption. Existing methods often satisfy only a subset of these three requirements.

Approaches based on diffractive optical elements (DOEs) \cite{Jeon2019,shi_2024_SnapHSI} or custom filter arrays \cite{Yako2023,Bian2024} require custom fabrication or permanent hardware modification. They can be compact and achieve high reconstruction performance, but their manufacturing cost is high. By contrast, methods using an RGB camera, with or without a single additional optical element (e.g., a prism) \cite{RGBHSI_Arad2016, RGBHSI_Nguyen2014,Survey_RGBHSI_2022, Baek2017, Fujimoto2025}, suffer from weak or spatially localized spectral modulation, which makes the inverse problem ill-conditioned. They can be low-cost and compact, but challenges remain in reconstruction accuracy or computational cost. Spatial-replication HSI \cite{Manakov_2013_SnapHSI_Kaleido,Daniel_2016_SnapHSI,Gorman:10_BRFSnapHSI} often enables simple and fast reconstruction, but image replication and spectral modulation are commonly implemented in separate stages, which increases system size due to relay optics or cascaded components.

To address this gap, we propose a spatial-replication snapshot HSI system that converts angular diversity into spectral-response diversity, as illustrated in \cref{fig:tether}.
A tapered kaleidoscope generates multiple replicated views with distinct incidence directions, and a directly attached birefringent filter (BRF) translates these directional differences into channel-dependent spectral transmittances.
Rather than treating the BRF's incidence-direction dependence as an error source, as is commonly done in conventional BRF-based imaging systems~\cite{Gorman:10_BRFSnapHSI,Kudenov:12_BRFSnap,salesin2022diy,Li:25}, we deliberately leverage it as a source of spectral-response diversity.
As a result, the replicated views provide complementary measurements that better condition the inverse problem for stable single-shot spectral reconstruction.
The proposed design also preserves a simple pixel-wise linear observation model, enabling fast and stable non-learning-based reconstruction.
It remains low-cost by relying on off-the-shelf optical components, and compact by integrating image replication and spectral modulation into a single attached unit that eliminates relay optics and cascaded modules.

\noindent The main contributions of this paper are as follows:
\begin{itemize}[
  label=\contribbullet,
]
\item
We propose angular-to-spectral diversity conversion for snapshot HSI by using kaleidoscope multi-view sampling and the incidence-direction dependence of a BRF, with a condition-number-based BRF selection criterion.
\item We present a low-cost (\(\sim\$1\mathrm{k}\)) and compact \SI{21}{\centi\metre} off-the-shelf kaleidoscope--BRF add-on that directly integrates spatial replication and BRF spectral modulation, eliminating relay stages.
\end{itemize}

\section{Related Work}
\label{sec:related_work}

\noindent\textbf{Snapshot HSI for Low-Cost and Compact Systems.}
Low-cost and compact snapshot HSI has been explored in several forms.
RGB-only approaches~\cite{RGBHSI_Arad2016, RGBHSI_Nguyen2014,Survey_RGBHSI_2022} are attractive for their simplicity, but their spectral responses are often limited and highly correlated, which can make the inverse problem ill-conditioned.
Adding a single coding element can improve measurement diversity. However, the resulting spectral cues often remain weak or spatially localized.
Baek \etal~\cite{Baek2017} combine an RGB camera with a prism, where spectral cues are sparse and reconstruction is solved iteratively with a reported runtime of about 45 minutes.
Zhao \etal~\cite{zhao2019hyperspectral} use a printer-generated random mask and reconstruct with a CNN trained on large paired data, with training reportedly taking about 24 hours on a single GPU.
Oh \etal~\cite{Oh2016_DIY} realize HSI using three cameras with different spectral sensitivity profiles, but this requires synchronized multi-camera capture and increases system size.

\noindent\textbf{Low-Cost Spatial-Replication Snapshot HSI.}
Spatial-replication snapshot HSI is closest to our approach.
We focus on low-cost designs, rather than higher-end alternatives~\cite{Gorman:10_BRFSnapHSI,HSI_spatialRep_Mu:19}.
Manakov \etal~\cite{Manakov_2013_SnapHSI_Kaleido} use a kaleidoscope purely as an image multiplier, then combine a filter array with relay optics so that each sub-image receives a distinct spectral modulation.
A key drawback is that the relay optics increase the system size. 
The overall system length is reported to be approximately \SI{100}{\centi \metre}~\cite{Manakov_2013_SnapHSI_Kaleido}.
Takatani \etal~\cite{takatani2017one} attach color filters directly onto mirror surfaces to simultaneously realize image replication and spectral multiplexing without relay optics.
However, the filter-on-mirror configuration can introduce diffuse reflections at the added interfaces, which may lead to blur in the replicated sub-images.
Compared with these systems, our method also uses spatial replication, but realizes spectral diversity through an attached BRF and incidence-direction differences, without relay optics or filter-on-mirror designs.

\noindent\textbf{BRF-Based Spectral Imaging.}
BRFs have been used for spectral imaging by varying the optical path difference (OPD) over time, for example by using liquid crystal controlled by an applied voltage~\cite{august2016miniature,Zhou2024_scanHSI,FAN2023_BRFScanHSIPol} or by mechanically rotating polarizers or birefringent elements~\cite{salesin2022diy}.
Such approaches obtain measurement diversity through time multiplexing and are therefore scanning HSI methods.
Snapshot HSI with birefringent optics includes cascaded polarization-splitting designs~\cite{Gorman:10_BRFSnapHSI}, in which increasing the number of channels generally requires a deeper cascade, as well as spatially varying OPD in natural materials~\cite{Fujimoto2025}, where spectral cues can remain localized.
Another line of work uses birefringent interferometry~\cite{Kudenov:12_BRFSnap}, which typically relies on specialized optics.
In contrast, our method obtains channel diversity from incidence-direction differences using a single attached BRF together with a kaleidoscope.

\noindent\textbf{Oblique-Incidence Effects in BRF.}
Because BRF transmittance varies with incidence direction~\cite{Veiras:10_BRF_OPD,Li:25,ZHANG2021_AngleN}, oblique incidence in imaging is often treated as a source of spectral non-uniformity.
In HSI, pixel-dependent ray angles arise naturally from the field of view and finite aperture cone, so many analyses assume near-normal incidence or adopt small-angle approximations.
Other studies explicitly model this effect or mitigate it by limiting the incidence cone or compensating wavelength shifts in post-processing~\cite{Li:25,Milind2021_angleN,ZHANG2021_AngleN}.
By contrast, in laser systems, BRFs are often operated under oblique incidence, where the incidence angle and optic axis parameters are deliberately chosen to control spectral selection and improve laser performance~\cite{wei2023_laser,Wei:21_laser,Demirbas:17_laser}.
Our method follows this latter perspective.
Rather than suppressing angle dependence, we exploit it as a source of diverse channel-wise spectral modulations for snapshot HSI.

A broader qualitative comparison with representative snapshot HSI systems is provided in \suppref{2}.
\section{A Pixel-Wise Linear Measurement Model for HSI}
\label{sec:why_measurement_diversity}

We use a standard pixel-wise linear model for snapshot HSI, assuming that all measurement channels are registered to the same scene point.
Let \(m \in \{1,\dots,M\}\) index the measurements, and let \(T_m(\lambda)\) denote the effective spectral response of channel \(m\).
The \(m\)-th measurement is modeled as
\begin{equation}
\label{eq:forward_overview_general}
I_m
=
\int_{\Lambda}
l(\lambda)\, T_m(\lambda)\, s(\lambda)\, d\lambda ,
\end{equation}
where \(l(\lambda)\) is the illumination spectrum and \(s(\lambda)\) is the spectral reflectance.

Discretizing \Cref{eq:forward_overview_general} over sampled wavelengths \(\{\lambda_k\}_{k=1}^{N_\lambda}\) gives
\begin{equation}
\label{eq:linear_model_general}
\mathbf{i}=\mathbf{H}\mathbf{s},
\end{equation}
where \(\mathbf{i}\in\mathbb{R}^{M}\) is the measurement vector, \(\mathbf{s}\in\mathbb{R}^{N_\lambda}\) is the spectral reflectance, and \(\mathbf{H}\in\mathbb{R}^{M\times N_\lambda}\) is the sensing matrix with
\begin{equation}
\label{eq:H_def_general}
H_{m,k}=l(\lambda_k)\,T_m(\lambda_k)\,\Delta\lambda_k .
\end{equation}

Reconstruction is governed by \(\mathbf{H}\), and its stability depends on how distinct the channel-wise spectral responses \(\{T_m(\lambda)\}_{m=1}^{M}\) are.

\section{Proposed System}
\label{sec:overview}
\begin{figure}[tb]
    \centering
    \begin{overpic}[width=\textwidth]{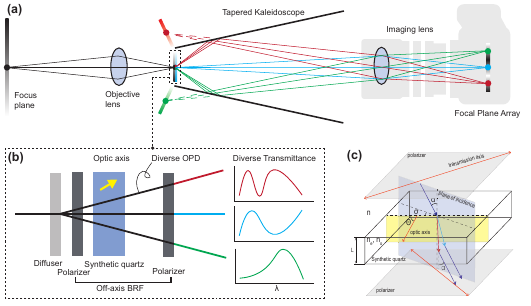}
        \put(8,94){\phantomsubcaption\label{fig:optics:a}}
        \put(49,94){\phantomsubcaption\label{fig:optics:b}}
        \put(84,94){\phantomsubcaption\label{fig:optics:c}}
    \end{overpic}
    \caption{Proposed optics: (a) System overview; (b) angular-to-spectral diversity conversion by the off-surface optic axis BRF; (c) angular geometry in the BRF.}
    \label{fig:optics}
\end{figure}

Building on the discussion in \cref{sec:why_measurement_diversity}, this section introduces the proposed system and explains how it converts angular diversity into spectral-response diversity.

\subsection{System Overview}
\label{ssec:System_Architecture}
To clarify the core idea behind our method, we begin with an overview of the optical system shown in \Cref{fig:optics:a,fig:optics:b}. The system consists of an objective lens, a diffuser, a BRF, a tapered kaleidoscope, and a camera.

The objective lens forms a 2D image of the scene on the diffuser plane.
The diffuser suppresses parallax induced by scene depth, ensuring that the replicated images share the same spatial structure across viewpoints\cite{Manakov_2013_SnapHSI_Kaleido,Daniel_2016_SnapHSI}. 
However, the diffuser’s Bidirectional Transmittance Distribution Function (BTDF) introduces differences in the spectral intensity distribution. This issue is addressed through calibration in our method.

A BRF is placed immediately behind the diffuser.
The BRF is mounted at the entrance of the tapered kaleidoscope.
The tapered kaleidoscope is a hollow mirrored tube that replicates the input image into multiple sub-images through internal reflections \cite{Han2003_taperd}.
These replicated sub-images are recorded by a single camera.
\subsection{Angular-to-Spectral Diversity Conversion}
\label{ssec:angular_to_spectral}

The central mechanism of the proposed system is angular-to-spectral diversity conversion.
The system first generates diversity in incidence directions across view channels, and then converts that angular diversity into spectral-modulation diversity through the incidence-direction-dependent spectral modulation of the BRF.

The tapered kaleidoscope produces multiple replicated sub-images, which we call view channels.
Because these channels correspond to different reflection paths, they also correspond to different incidence directions at the BRF.
Let \((\alpha_v,\sigma_v)\) denote the incidence direction associated with view channel \(v \in \{1,\dots,V\}\), where \(\alpha_v\) is the incidence angle and \(\sigma_v\) is the azimuth (\Cref{fig:optics:c}).
The taper enlarges the spread of these sampled directions, increasing the angular diversity across channels.

This angular diversity is then converted into channel-wise spectral-response diversity by the BRF.
A BRF is implemented as a birefringent plate sandwiched between two linear polarizers.
After passing through the first polarizer, the incident light becomes linearly polarized and splits inside the birefringent plate into ordinary and extraordinary components.
These components accumulate an OPD, denoted by \(\Delta_{o-e}\), and the second polarizer projects both components onto a common polarization axis, resulting in a transmitted intensity that varies with $\Delta_{o-e}/\lambda$.
Because \(\Delta_{o-e}\) depends on the incidence direction, the BRF spectral transmittance also changes with \((\alpha,\sigma)\).
Under the crossed-Nicols configuration shown in \Cref{fig:optics:c}, the BRF spectral transmittance is written as
\begin{equation}
\label{eq:brf_transmittance}
B(\lambda;\alpha,\sigma,\theta)
\propto
\sin^2\!\left(
\frac{\pi\,\Delta_{o-e}(\alpha,\sigma,\theta)}{\lambda}
\right),
\end{equation}
where \(\theta\) denotes the optic axis diving angle (the angle between the optic axis and the surface plane).
Therefore, the different incidence directions \(\{(\alpha_v,\sigma_v)\}_{v=1}^{V}\) generated by the kaleidoscope are mapped to different BRF spectral transmittances,
\begin{equation}
\label{eq:Bv_def}
B_v(\lambda)=B(\lambda;\alpha_v,\sigma_v,\theta).
\end{equation}

The camera further observes each view through one of the RGB sensor sensitivities \(f^c(\lambda)\), where \(c \in \{R,G,B\}\).
Hence, the effective spectral response associated with view channel \(v\) and color channel \(c\) becomes
\begin{equation}
\label{eq:Tvc_def}
T_{v,c}(\lambda)=f^c(\lambda)\,B_v(\lambda).
\end{equation}
As a result, the angular diversity produced by the kaleidoscope is converted into channel-wise spectral-response diversity \(\{\{T_{v,c}(\lambda)\}_{v=1}^{V}\}_{c \in \{R,G,B\}}\), which provides the measurement diversity required for stable spectral reconstruction.

Thanks to the diffuser, the view channels retain nearly the same spatial structure and differ mainly in spectral modulation. This enables snapshot spectral multiplexing with a simple channel-wise image formation model. The kaleidoscope produces angular diversity, and the BRF converts it into spectral-response diversity across channels, yielding complementary measurements that improve the conditioning of the inverse problem. Because the BRF is mounted directly at the entrance of the tapered kaleidoscope, this angular-to-spectral conversion is realized within a compact optical architecture, without relay optics or cascaded components.

\subsection{Pixel-Wise Spectral Reconstruction}
\label{ssec:Spectral_Reconstruction}

Using the effective spectral response \(T_{v,c}(\lambda)\) defined in \Cref{eq:Tvc_def}, the sensing matrix of the proposed system is given by
\begin{equation}
\label{eq:H_def_ours}
H_{(v,c),k}
=
l(\lambda_k)\,f^c(\lambda_k)\,B_v(\lambda_k)\,\Delta\lambda_k.
\end{equation}

After inter-channel registration, we reconstruct the spectral reflectance at each pixel by solving a non-negative least-squares problem with second-order spectral smoothness:
\begin{equation}
\label{eq:argmin_overview}
\hat{\mathbf{s}}(\mathbf{x})
=
\arg\min_{\mathbf{s}\ge 0}
\;
\|\mathbf{H}\mathbf{s}-\mathbf{i}(\mathbf{x})\|_2^2
+
\beta
\|\mathbf{D}\mathbf{s}\|_2^2,
\end{equation}
where \(\mathbf{D}\) is the second-order difference matrix along wavelength, and \(\beta\) is the smoothness weight.

Because the proposed system provides sufficient measurement diversity through diverse channel-wise spectral responses while preserving a simple spatial-replication image formation model, spectral reconstruction reduces to a lightweight regularized linear inverse problem at each pixel, without learning.

\section{BRF Design and Parameter Selection}
\label{sec:BRF-design}
\begin{figure}[t]
    \centering
    \includegraphics[width=\linewidth]{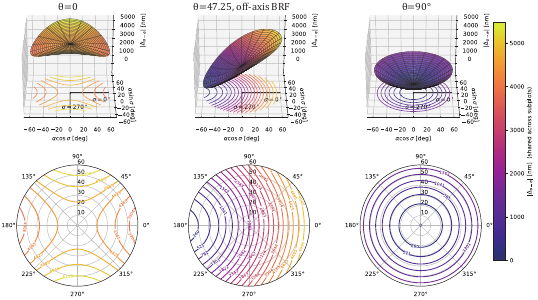}
\caption{
    Simulated OPD magnitude as a function of incidence direction $(\alpha,\sigma)$.
    Top: 3D surface with $(x,y)=(\alpha\cos\sigma,\alpha\sin\sigma)$. Bottom: polar contour with radial coordinate $\alpha$ and azimuth $\sigma$.
    Parameters: $n=1$, $n_o=1.54264$, $n_e=1.55170$, $L=5\times10^{-4}$\,m.
    }   
   
    \label{fig:OPD_map}
\end{figure}

This section presents a BRF design procedure for the proposed system, using a conditioning-based criterion to select a suitable off-the-shelf configuration.

\subsection{OPD-Based Model Under Oblique Incidence}
\label{ssec:opd_model}

In the proposed system, the tapered kaleidoscope determines a discrete set of incidence directions \(\{(\alpha_v,\sigma_v)\}_{v=1}^{V}\) at the BRF.
For a birefringent plate under oblique incidence with a diving optic axis, the OPD can be written as follows \cite{Veiras:10_BRF_OPD}:
\begin{equation}
\label{eq:OPD}
\begin{aligned}
\Delta_{o-e}
&= L\Bigl[
\sqrt{n_o^2-(n\sin\alpha)^2}
+ \frac{n\,(n_o^2-n_e^2)\sin\theta\cos\theta\cos\sigma\sin\alpha}{D}
\\
&\qquad
- \frac{n_o}{D}
\sqrt{
n_e^2 D
-
\Bigl(n_e^2-(n_e^2-n_o^2)\cos^2\theta\sin^2\sigma\Bigr)(n\sin\alpha)^2
}
\Bigr],
\\
&\text{where } D = n_e^2\sin^2\theta+n_o^2\cos^2\theta .
\end{aligned}
\end{equation}
Here, \(n\) is the refractive index of the surrounding isotropic medium, \(n_o\) and \(n_e\) are the ordinary and extraordinary refractive indices of the birefringent plate, and \(L\) is the plate thickness.

Combined with \Cref{eq:brf_transmittance}, \Cref{eq:OPD} determines how the BRF spectral transmittance varies across the sampled incidence directions.
Accordingly, the BRF design problem is to choose \(\theta\) such that the resulting set of channel-wise transmittances \(\{B_v(\lambda)\}_{v=1}^{V}\) provides strong spectral-response diversity.

\subsection{Design Implications of OPD Symmetry}
\label{ssec:opd_symmetry}

A large variation of \(\Delta_{o-e}\) over \((\alpha,\sigma)\) is helpful, but it is not sufficient by itself.
What matters is whether the sampled directions \(\{(\alpha_v,\sigma_v)\}_{v=1}^{V}\) map to distinct OPD values.
If multiple view channels yield similar OPDs, their BRF transmittances become similar, the corresponding measurements are redundant, and the effective spectral responses in \(\mathbf{H}\) become more correlated.

This issue is particularly important for spatial-replication snapshot HSI, where the sampled directions are constrained by the kaleidoscope geometry and often contain structured azimuthal relationships.
As an extreme case, when $\theta=$~\ang{90}, \Cref{eq:OPD} becomes independent of \(\sigma\), so the BRF provides no azimuthal diversity.

In the conventional on-surface optic axis BRF configuration ($\theta=$~\ang{0}), the OPD is invariant under a \(180^\circ\) azimuthal rotation,
\begin{equation}
\label{eq:opd_centrosym_on}
\Delta_{o-e}(\alpha,\sigma;0)=\Delta_{o-e}(\alpha,\sigma+\pi;0),
\end{equation}
which causes channel pairs under symmetric angular sampling to collapse to the same OPD.
This reduces measurement diversity by making the channel-wise spectral modulations more similar and can make the inverse problem ill-conditioned.

By contrast, for an off-surface optic axis BRF $(0<\theta<\ang{90})$, the term proportional to $\cos\sigma\,\sin\alpha$ in \Cref{eq:OPD} breaks the $\sigma\to\sigma+\pi$ invariance.
As shown in \Cref{fig:OPD_map}, this removes the half-turn repetition observed in the on-surface optic axis case and reduces redundancy among the sampled view channels.
A mirror symmetry still remains,
\begin{equation}
\label{eq:opd_symmetry_off}
\Delta_{o-e}(\alpha,\sigma;\theta)=\Delta_{o-e}(\alpha,-\sigma;\theta),
\end{equation}
but this pairing is typically less harmful than the \(\sigma\to\sigma+\pi\) equivalence under rotationally structured sampling.

Moreover, \Cref{fig:OPD_map} shows that off-surface optic axis settings also increase the spread of \(|\Delta_{o-e}|\) over \((\alpha,\sigma)\) compared with the on-surface optic axis case.
This broader OPD distribution strengthens the differences among the channel-wise BRF transmittances in \Cref{eq:brf_transmittance}.
Together, these observations motivate the use of an off-surface optic axis BRF in the proposed system.

\subsection{Design Criterion: Conditioning of the Sensing Matrix}
\label{ssec:Design_Criterion}

The remaining question is which off-surface optic axis configuration best improves spectral reconstruction.
Because the effective spectral responses are encoded in the sensing matrix \(\mathbf{H}\) in \Cref{eq:H_def_ours}, we evaluate BRF designs through the conditioning of \(\mathbf{H}\).

Specifically, we use the condition number
\begin{equation}
\label{eq:cond_def_simple}
\kappa(\mathbf{H}) \coloneqq
\frac{\sigma_{\max}(\mathbf{H})}{\sigma_{\min}(\mathbf{H})},
\end{equation}
where \(\sigma_{\max}\) and \(\sigma_{\min}\) are the largest and smallest singular values of \(\mathbf{H}\), respectively.
When \(\sigma_{\min}\) is small, \(\mathbf{H}\) becomes nearly rank-deficient, the inverse problem becomes unstable, and stronger regularization is required.
Therefore, a smaller \(\kappa(\mathbf{H})\) indicates that the channel-wise spectral modulations are more complementary and that reconstruction is more stable.

In our setting, the incidence-direction samples \(\{(\alpha_v,\sigma_v)\}_{v=1}^{V}\) are determined by the tapered-kaleidoscope geometry, while \(l(\lambda)\) and the camera sensitivities \(f^c(\lambda)\) are obtained by calibration.
Hence, once the sampled directions are fixed, we can precompute \(\mathbf{H}\) for each candidate BRF configuration and use \(\kappa(\mathbf{H})\) as a practical design metric.
This gives a direct parameter-selection procedure: determine the angular samples produced by the kaleidoscope, evaluate candidate BRF settings through \(\kappa(\mathbf{H})\), and choose the configuration that yields the best conditioning.
We next apply this procedure to the physical parameter choices used in our prototype.

An experimental validation of the OPD distribution using measured transmittance profiles is provided in \suppref{4}.

\subsection{Applying the Criterion to Physical Parameter Selection}
\label{ssec:parameter_selection}

We first fix the tapered-kaleidoscope geometry, which defines the discrete incidence-direction samples at the BRF, and then select the BRF configuration that best conditions the sensing matrix under those samples.

\noindent\textbf{Tapered kaleidoscope.}
We use a tapered kaleidoscope to obtain a diverse set of incidence directions at the BRF.
Following \cite{Han2003_taperd}, we adopt a 3-sided design and set the taper angle to \(9^\circ\), which reduces view-channel fragmentation on the sensor while maintaining sharp and stable multi-reflection views.

Because tapering makes the replicated virtual images behave as samples on a virtual viewing sphere, different reflection orders can have different apparent distances.
To keep all view channels simultaneously in focus without requiring an excessively small aperture, we use the \(V=10\) view channels produced by reflection orders up to two.
Given this kaleidoscope geometry, we compute the sampled incidence directions \(\{(\alpha_v,\sigma_v)\}_{v=1}^{V}\) by ray tracing, and we also scan the in-plane mounting rotation \(b\), which shifts the azimuths as \(\sigma_v \leftarrow \sigma_v + b\).
These discrete angular samples are then used in the conditioning-based BRF selection below.

\begin{figure}[t]
  \centering
  \begin{subfigure}[t]{0.55\linewidth}
    \centering
    \includegraphics[width=\linewidth]{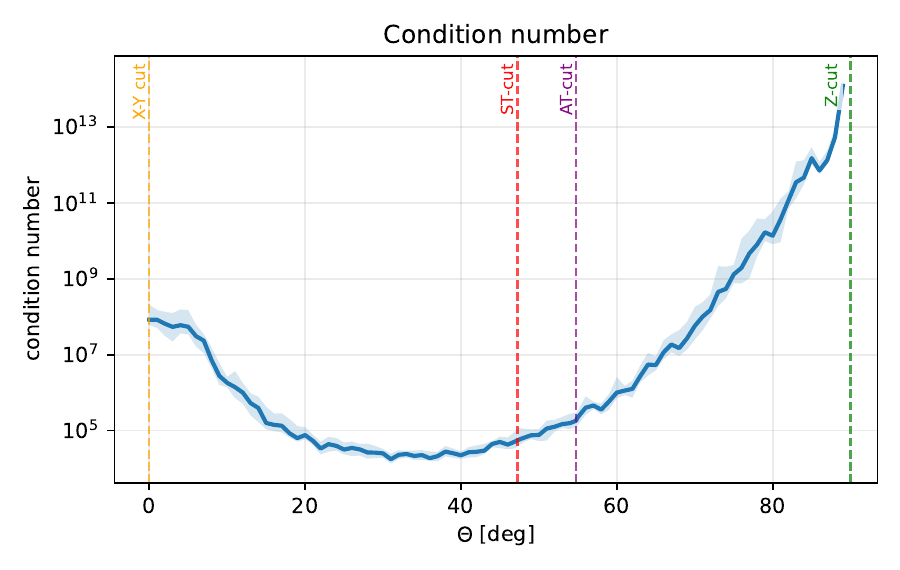}
    \caption{Condition number $\kappa(\mathbf{H}(\theta))$ vs.\ $\theta$ over in-plane mounting rotation $b$.}
    \label{subfig:Condition_number_theta}

  \end{subfigure}\hfill
  \begin{subfigure}[t]{0.43\linewidth}
    \centering
    \includegraphics[width=\linewidth]{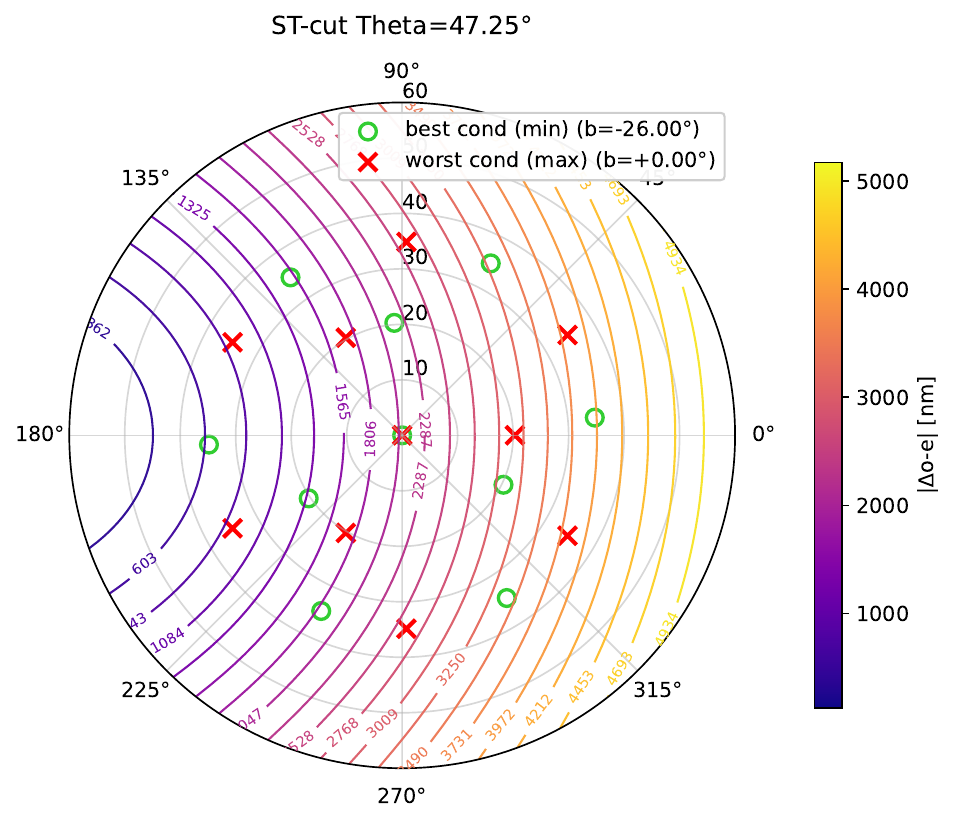}
    \caption{OPD contour plot of an ST-cut quartz plate, with sampled incidence directions.}
    \label{subfig:OPD_map}
  \end{subfigure}
\caption{Optic axis selection using conditioning.}  \label{fig:theta_choice_cond}
\end{figure}

\noindent\textbf{BRF.}
Given the sampled incidence directions \(\{(\alpha_v,\sigma_v)\}_{v=1}^{V}\) obtained above, we select the BRF by minimizing the condition number introduced in \Cref{ssec:Design_Criterion}.
For each candidate optic axis diving angle \(\theta\), we scan the in-plane mounting rotation \(b\) and evaluate the resulting \(\kappa(\mathbf{H})\). 
\Cref{subfig:Condition_number_theta} summarizes the results.

We use a synthetic quartz plate as the birefringent plate in the BRF.
Synthetic quartz is widely used in frequency-control resonators and is commercially available in standard cuts with known optic axis angles.
This makes it a practical low-cost choice that is consistent with our off-the-shelf design goal.
We therefore sweep these cut-defined candidates and select the one that best conditions \(\mathbf{H}\).
Among these candidates, an ST-cut synthetic quartz plate ($\theta=$~\ang{47.25}) yields the best conditioning.
We thus use an ST-cut synthetic quartz plate (MTI Corporation, \SI{10}{\milli \metre} \(\times\) \SI{10}{\milli \metre} \(\times\) \SI{0.5}{\milli \metre}) 
and assemble a BRF in a crossed-Nicols configuration using two linear film polarizers.

For the in-plane mounting rotation \(b\), the condition number is exceptionally large at $b=$~\ang{0} (\(1.66\times10^{16}\)), indicating that \(\mathbf{H}\) is effectively rank-deficient.
This is consistent with the residual symmetry of the off-surface optic axis OPD map (\cref{subfig:OPD_map}).
At this rotation, multiple sampled incidence directions fall onto nearly identical OPD values, which reduces spectral-modulation diversity.
We therefore avoid $b=$~\ang{0} and assemble the system at $b=$~\ang{20}.

\section{Implementation Details}
\label{sec:Prototype}
\subsection{Other Hardware Setup}
\label{ssec:hardware_setup}

\noindent\textbf{Diffuser.}
We use a wide-angle holographic diffuser (Edmund Optics, \ang{80} Diffusing Angle, 2 inch square) to support the wide range of  incidence angles in our multi-view observation.
Its transmission drops to about \SI{50}{\percent} at an incidence angle of \ang{40}.
Given our maximum incidence angle of \ang{34.86}, this angular roll-off can reduce the optical throughput of high-reflection view channels.

\noindent\textbf{Other physical dimensions.}
The prototype is a \SI{21}{\centi\metre} add-on unit (approx. \SI{40}{\centi\metre} including the camera) built with \SI{35}{\milli\metre} objective lens (Edmund Optics, C-mount)  and a camera (Sony $\alpha$7R IV) with a zoom lens (TAMRON, at the telephoto end: \SI{40}{\milli\metre}).
The camera is placed approximately \SI{30}{\centi\metre} from the BRF, constrained by the lens minimum focus distance.

\noindent\textbf{Add-on component parts cost.}
The approximate costs of the add-on components are: polarizers \(\approx\)\$8, mirrors \(\approx\)\$20 total (custom-cut front-surface mirrors readily ordered from a consumer-facing retailer), synthetic quartz \$33.35, diffuser \$273, and objective lens \$635.
Overall, the deployed add-on hardware remains within \$1k, excluding the camera body, imaging lens, and calibration equipment such as a monochromator.

\subsection{Preprocessing}
\label{sec:calibration}

To construct the sensing matrix \(\mathbf{H}\) in \Cref{eq:H_def_ours}, we pre-estimate
(i) the view-channel-wise BRF transmittance \(B_v(\lambda)\),
(ii) the product of the illumination spectrum and camera RGB spectral sensitivity, \(\phi^c(\lambda)=l(\lambda)f^c(\lambda)\),
and wavelength-independent throughput factors \(g_v\).

\noindent\textbf{(i) Measuring \(B_v(\lambda)\).}
We measure \(B_v(\lambda)\) using monochromator-based narrowband illumination and estimate the channel-wise transmittance profiles from the captured images.
This calibration absorbs the diffuser BTDF into the estimated \(B_v(\lambda)\).
To account for slight within-channel variation, we use pixel-wise calibrated transmittance profiles when constructing the per-pixel sensing matrix.

\noindent\textbf{(ii) Estimating \(\phi^c(\lambda)\) and throughput factors \(g_v\).}
Assuming a spectrally flat reflectance for a white reference target, the observation of view channel \(v\) and color channel \(c\) under the white target can be written as
\begin{equation}
I_{v,\mathrm{w}}^{c}
\approx
g_v
\int_{\Lambda}
\phi^c(\lambda)\,B_v(\lambda)\,d\lambda ,
\label{eq:calib_white}
\end{equation}
where \(g_v\) accounts for channel-wise throughput differences.
We discretize \Cref{eq:calib_white} and estimate \(\phi^c(\lambda)\) and \(g_v\) jointly via regularized least squares.
We then apply channel-wise gain correction and construct \(\mathbf{H}\) from the calibrated quantities according to \Cref{eq:H_def_ours}.

\noindent\textbf{Inter-channel registration.}
We align view channels using a calibration target and estimate inter-channel homographies, including mirror-induced parity flips.
The estimated registration parameters, \(\mathbf{H}\), and \(g_v\) are fixed in subsequent reconstruction.

Full calibration details are provided in \suppref{1}.

\section{Experiments}

\subsection{Experiments on Synthetic Data}
\label{sec:exp_synth}

\begin{figure}[tb]
    \centering
    \begin{subfigure}[t]{0.5\linewidth}
        \centering
        \includegraphics[width=\linewidth]{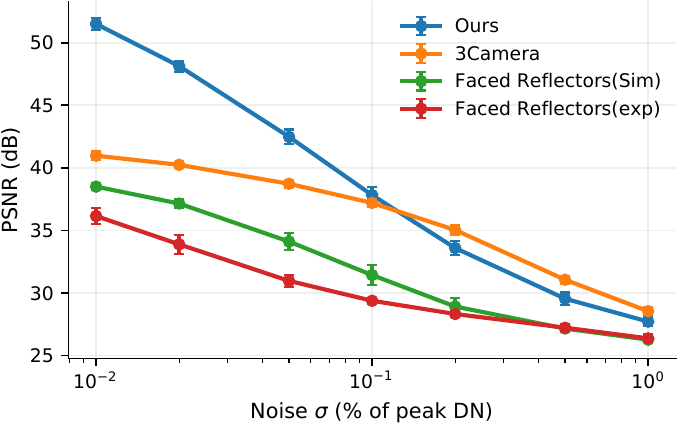}
        \caption{PSNR vs.\ additive noise (exposure-matched).}
        \label{fig:psnr_noise_curve}
    \end{subfigure}
    \hfill
    \begin{subfigure}[t]{0.48\linewidth}
        \centering
        \includegraphics[width=\linewidth]{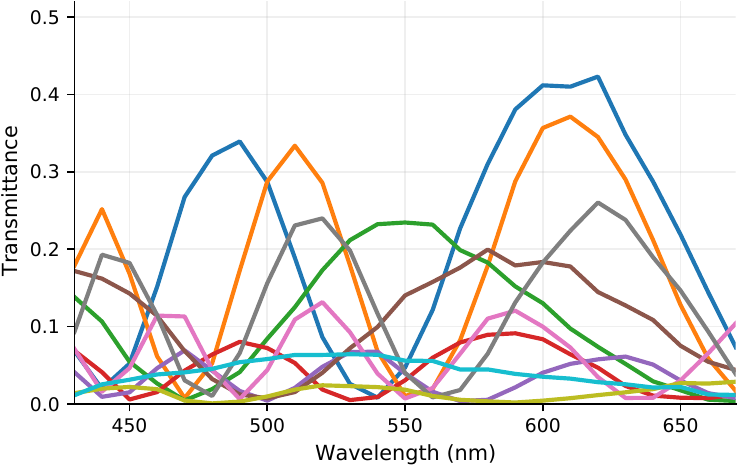}
        \caption{Measured transmittance of our system.}
        \label{fig:Transmittances}
    \end{subfigure}

    \caption{
    Synthetic ColorChecker comparison with low-cost snapshot baselines.
    }    \label{fig:synth_psnr_vs_noise_and_ss}
\end{figure}

\noindent\textbf{Experimental Setup.}
We evaluate our system against two low-cost snapshot baselines, \textit{3Camera}~\cite{Oh2016_DIY} and \textit{Faced Reflectors}~\cite{takatani2017one}, focusing on the measurement diversity induced by our channel-wise spectral-response diversity.
To isolate the effect of the optics, we reconstruct spectra for all methods using the same non-negative, smoothness-regularized least-squares formulation in \Cref{eq:argmin_overview}.
We generate synthetic observations from the 24 ColorChecker reflectance spectra using each method's image formulation model, and add i.i.d.\ Gaussian noise after exposure matching such that the maximum DN is approximately 255.
The measured transmittance profiles used in our method are shown in \Cref{fig:Transmittances}.
For each noise level, we sweep the smoothness weight \(\beta\) and report the best PSNR over \(\beta\), as shown in \Cref{fig:psnr_noise_curve}.

See \suppref{3} for details.

\noindent\textbf{Result Analysis.}
Our method achieves higher PSNR than the baselines in the low-noise regime, suggesting that the proposed system provides richer spectral-response diversity across channels and hence greater measurement diversity for the inverse problem.
As noise increases, the PSNR gap narrows and can slightly reverse in the high-noise regime.
This behavior may be explained by non-uniform optical throughput across view channels.
Because our channels differ not only in spectral transmittance but also in overall optical throughput, the resulting SNR is imbalanced; when noise dominates, low-throughput channels contribute less as effective measurements, diminishing the benefit of additional channels.
This throughput imbalance is amplified by angle-dependent diffuser emission and by the polarizer loss in the BRF.

\subsection{Experiments on Real Data}
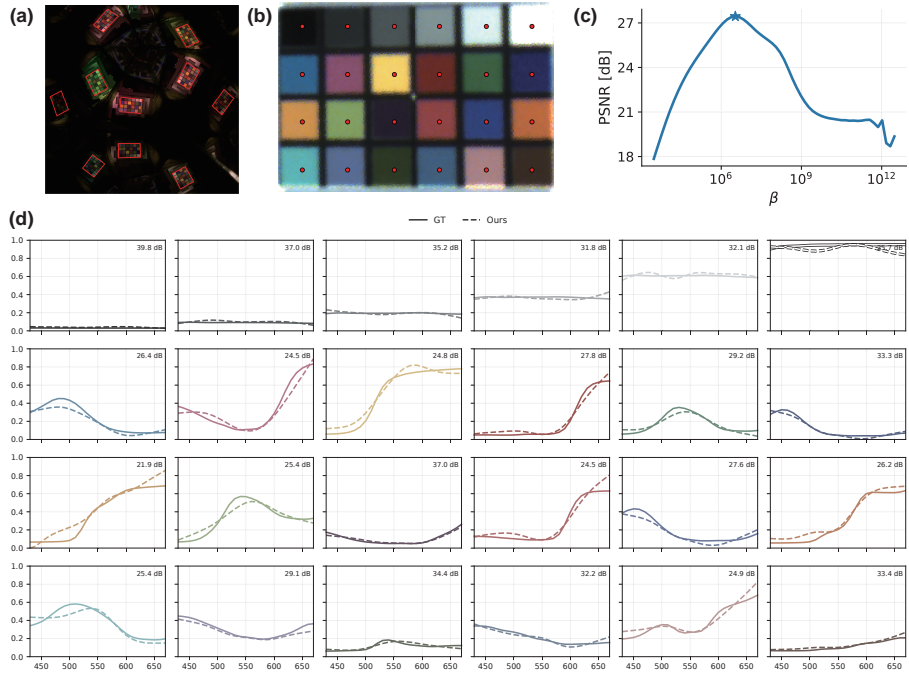
\begin{figure}[tb]
    \centering
    \begin{overpic}[width=\textwidth]{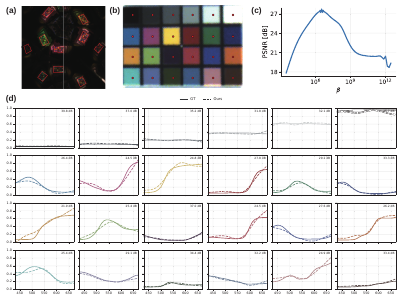}
        \put(8,94){\phantomsubcaption\label{fig:result_CCheck:a}}
        \put(49,94){\phantomsubcaption\label{fig:result_CCheck:b}}
        \put(84,94){\phantomsubcaption\label{fig:result_CCheck:c}}
        \put(86,94){\phantomsubcaption\label{fig:result_CCheck:d}}
    \end{overpic}
    \caption{
    ColorChecker reflectance reconstruction: (a) capture crop; (b) RGB rendering; red dots mark the patch sampling locations; (c) PSNR vs.\ smoothness weight $\beta$; (d) reconstructed and reference spectra.
        }
    \label{fig:result_CCheck}
\end{figure}

\noindent\textbf{ColorChecker Validation.}
We calibrate the illumination spectrum and camera RGB sensitivities using a gray balance target, and reconstruct spectral reflectance of an X-Rite ColorChecker Passport Photo 2 from the 10 view channels in \Cref{fig:result_CCheck:a}. 
We illuminate the scene with an incandescent lamp.
We reconstruct 25 bands over \qtyrange{430}{670}{\nano\metre} at \SI{10}{\nano\metre} intervals. 
We used conservative settings (ISO 125, F/8, \SI{8}{\second}) for robust focus and SNR across view channels.
Since the recovered reflectance is defined up to a global scale, we fit a single scalar using all 24 patches by least squares against the reference spectra and apply it to all reconstructions. 
We determine a single smoothness weight $\beta$ from the ColorChecker experiment (\Cref{fig:result_CCheck:c}) and keep it fixed for the real-scene and USAF-target reconstructions.
At this $\beta$, \Cref{fig:result_CCheck:b,fig:result_CCheck:d} show the RGB rendering and reconstructed spectra.
The RGB rendering is obtained by projecting the reconstructed reflectance cube onto the calibrated camera RGB sensitivities \(f^c(\lambda)\), \(c\in\{R,G,B\}\). 
Each rendered channel is normalized by the 95th percentile of valid pixels, and no display gamma correction is applied.
The PSNR computed over all 24 patches and all wavelengths is \SI{27.4}{\decibel}, confirming that our system enables reasonably accurate spectral reconstruction.

\noindent\emph{Runtime.} We measure per-frame reconstruction time for 25-band reflectance on $226\times155$ pixels, averaged over five RAW frames, on a workstation with an NVIDIA Quadro RTX 8000 GPU.
The total runtime is \SI{0.748}{\second}/frame, of which solving \Cref{eq:argmin_overview} takes \SI{0.044}{\second}/frame (\(\approx\)\SI{5.9}{\percent}), while most time is spent on RAW loading and forming the observation vector $\mathbf{i}$.
A one-time preprocessing step for illumination and camera-sensitivity calibration takes \(\sim\)\SI{26}{\second} and is reused as long as the configuration is unchanged. 

\begin{figure}[tb]
\centering
\includegraphics[width=\linewidth]{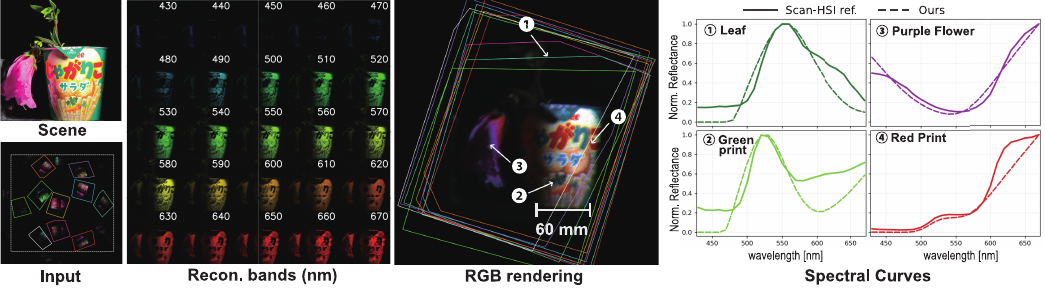}
\caption{
Real-scene 25-band reconstruction.
Colored outlines indicate registered view-channel regions; the white dashed box is used for the valid-area ratio.
}
\label{fig:real_scene_exp}
\end{figure}

\noindent\textbf{Real-Scene Validation.}
To evaluate the prototype beyond flat ColorChecker patches, we captured a real scene containing a leaf, a flower, and printed materials under daylight-level artificial solar illumination (SERIC SOLAX XC-100).
The scene was captured using the same prototype with a \SI{0.25}{\second} exposure, ISO~200, and F/8.
We reconstructed 25 bands over \qtyrange{430}{670}{\nano\metre} using the same calibration, inter-channel registration, reconstruction pipeline, and the $\beta$ selected in the ColorChecker validation.
As shown in \Cref{fig:real_scene_exp}, the reconstructed spectral bands and point spectra exhibit material-dependent spectral variations across the scene.
Since the commercial scanning-HSI reference was captured from a different viewpoint, it is not pixel-aligned with our reconstruction and is used only to compare spectral trends at the marked points.
Residual errors remain near object boundaries and printed-text edges, indicating the practical limits of registration accuracy and diffuser-based parallax suppression.

\noindent\textbf{Spatial Resolution.}
To quantify the spatial-resolution trade-off of the 10-view configuration, we evaluated a USAF target.
Full details are provided in \suppref{5}.
Most reconstructed HSI bands show resolvable structures at \SIrange{0.89}{1.00}{\linepair\per\milli\metre}, compared with \SI{1.12}{\linepair\per\milli\metre} for an RGB capture without the kaleidoscope/diffuser under the same objective lens, camera, and image-plane scale.
Thus, the limiting resolution is within one to two USAF group/element steps of the RGB capture without the kaleidoscope/diffuser.
The main spatial cost of the current prototype is instead the reduced valid reconstruction area caused by 10-view replication and the non-optimized camera--kaleidoscope field-of-view matching.
The valid region contains approximately \(3.24\times10^{5}\) pixels, or \SI{4.5}{\percent} of the dashed box enclosing the view-channel regions in \Cref{fig:real_scene_exp}.

\subsection{Ablation Study}
\begin{figure}[t]
  \centering

  \begin{subfigure}[t]{0.32\linewidth}
    \centering
    \includegraphics[width=\linewidth]{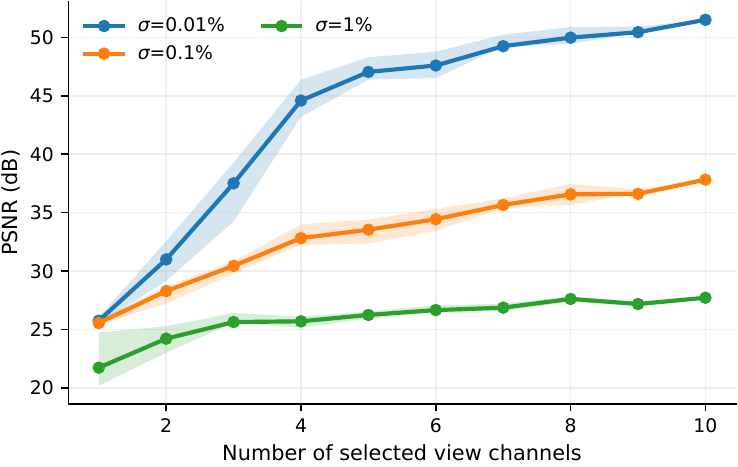}
    \caption{PSNR vs.\ \#view channels}
    \label{fig:psnr_vs_filter}
  \end{subfigure}
  \hfill
  \begin{subfigure}[t]{0.36\linewidth}
    \centering
    \includegraphics[width=\linewidth]{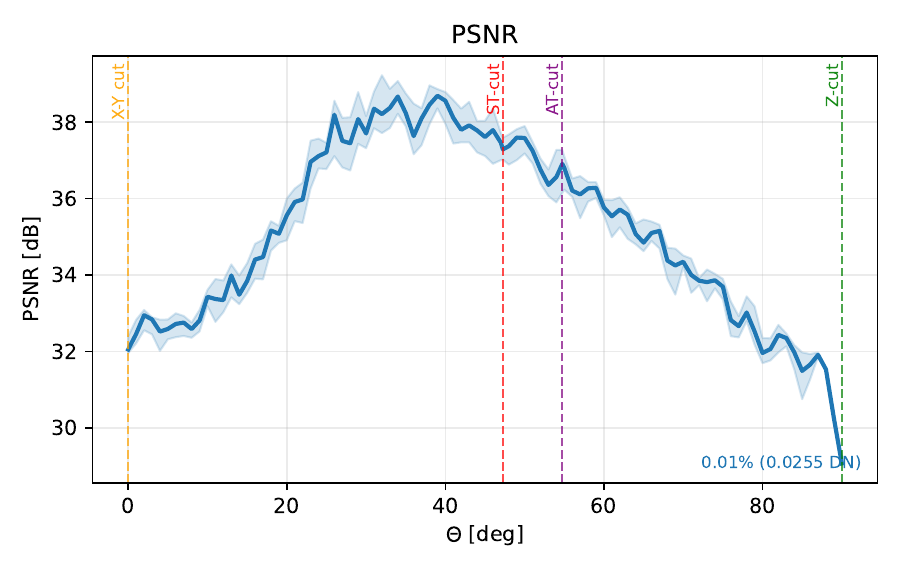}
    \caption{PSNR vs.\ $\theta$}
    \label{fig:theta_psnr}
  \end{subfigure}
  \hfill
  \begin{subfigure}[t]{0.30\linewidth}
    \centering
    \includegraphics[width=\linewidth]{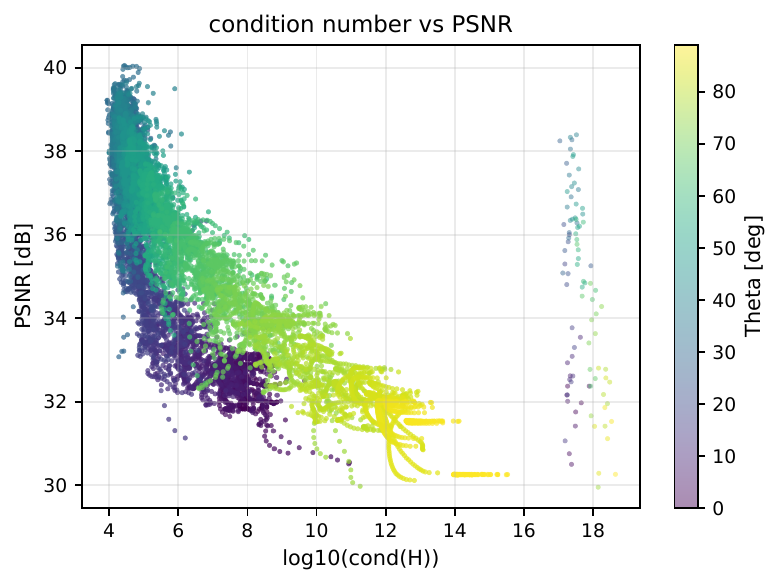}
    \caption{$\kappa(\mathbf{H})$--PSNR correlation}
    \label{fig:cond_psnr_corrmap}
  \end{subfigure}
    \caption{Synthetic ColorChecker ablation study.}
    \label{fig:ablation_summary}
\end{figure}

We ablate the design factors that determine measurement diversity under the same synthetic protocol as in \Cref{sec:exp_synth}. The goal is to test whether reconstruction accuracy follows the conditioning-based design analysis in \Cref{sec:BRF-design}.

\noindent\textbf{Effect of the Number of View Channels.}
We vary the number of used view channels from $V=10$ down to $1$.
PSNR decreases monotonically as $V$ decreases (\Cref{fig:psnr_vs_filter}),
showing that the view channels provide non-redundant spectral constraints for reconstruction.

\noindent\textbf{Effect of BRF Angle Configuration.}
We vary the optic axis angle $\theta$ and the mounting rotation $b$, and evaluate PSNR with the same reconstructor.
As shown in \Cref{fig:theta_psnr}, on-surface optic axis or poorly aligned settings are consistently worse than the selected off-surface optic axis setting, indicating that the BRF angle configuration directly affects usable spectral diversity.

\noindent\textbf{Relationship Between Conditioning and PSNR.}
\Cref{fig:cond_psnr_corrmap} shows that PSNR follows the trend of $\kappa(\mathbf{H})$ across $(\theta,b)$, supporting condition-number minimization as a practical design rule.

\section{Conclusion}
We presented a compact, low-cost spatial-replication snapshot HSI system that uses a directly attached tapered kaleidoscope--BRF unit to convert kaleidoscope-induced angular diversity into BRF-based spectral-response diversity.
The proposed design preserves a pixel-wise linear measurement model, enables fast non-learning-based spectral reconstruction, and selects the BRF configuration using a condition-number-based criterion.
Experiments on synthetic data, ColorChecker validation, and real-scene reconstruction with an off-the-shelf prototype validate stable and reasonably accurate 25-band reconstruction in a compact add-on form factor.

\noindent\textbf{Limitations.}
The prototype exhibits channel-dependent throughput due to diffuser angular roll-off, which reduces the effective contribution of some view channels in noise-dominated regimes and leads to long exposure under conservative ColorChecker settings.
Its effective spatial resolution is limited by dividing the sensor area among replicated views.
The current implementation also requires one-time monochromator-based calibration outside the add-on component cost; reducing this burden through model-assisted calibration is future work.

\subsubsection*{Acknowledgements}
This work was supported by JST SPRING JPMJSP2140, JSPS KAKENHI JP23H00499, and JST CREST JPMJCR23N3.

%
%
\bibliographystyle{splncs04}
\bibliography{main}
\end{document}


\title{Compact Low-Cost Hyperspectral Imaging\texorpdfstring{\\}{ }via Angular-to-Spectral Diversity Conversion\texorpdfstring{\\}{ }--- Supplementary Material ---} 

\titlerunning{Compact Low-Cost HSI via Angular-to-Spectral Diversity Conversion}

\newcommand\correspondingfootnote{%
  \begingroup
  \renewcommand{\thefootnote}{\dag}%
  \renewcommand{\theHfootnote}{correspondingauthor}%
  \begin{NoHyper}
  \footnotetext{Corresponding author.}%
  \end{NoHyper}
  \endgroup
}\author{Kazuma Fujiwara\textsuperscript{\dag}\inst{1}\orcidlink{0009-0007-2989-5169} \and
Takuya Funatomi\inst{1,2}\orcidlink{0000-0001-5588-5932} \and
Kazuya Kitano\inst{1}\orcidlink{0000-0001-6430-6963}
\and 
Yuki~Fujimura\inst{1}\orcidlink{0000-0002-7225-8452}
\and
Yasuhiro Mukaigawa\inst{1}\orcidlink{0000-0001-8689-3724}
}

\authorrunning{K.~Fujiwara et al.}

\institute{
Nara Institute of Science and Technology, Nara, Japan
\and
Kyoto University, Kyoto, Japan\\
\email{\{fujiwara.kazuma.fj4, kitano.kazuya, fujimura.yuki, mukaigawa\}@is.naist.jp}, 
\email{funatomi.takuya.2c@kyoto-u.ac.jp}
}

\maketitle
\correspondingfootnote

In this supplementary material, we provide additional methodological and experimental details that support the main text.
We first present the full preprocessing procedure, then provide an additional qualitative comparison with representative snapshot HSI systems, describe the detailed synthetic experimental setup, provide an experimental validation of the optical path difference (OPD) distribution using real data, and finally report the spatial-resolution evaluation using a USAF target.

\section{Full Preprocessing Details}
\begin{figure}[t]
    \centering
    \includegraphics[width=0.6\linewidth]{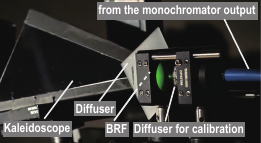}
    \caption{
    Calibration setup for measuring the view-channel-wise BRF transmittance $B_v(\lambda)$.
    }
    \label{fig:calib_setup}
\end{figure}

\subsection{Measuring View-Channel-Wise Birefringent Filter Transmittance}

We calibrate the view-channel-wise effective birefringent filter (BRF) transmittance $B_v(\lambda)$ using the setup shown in \Cref{fig:calib_setup}.
A monochromator (Optometrics, DMC1-03) equipped with a tungsten halogen light source generates narrowband illumination.
Specifically, we use the tungsten halogen source (Optometrics, TS-110), which mounts directly to the monochromator.
The monochromator output is split by a bifurcated fiber into two paths: one path is directed to our optical system, and the other is directed to a spectrometer.
The spectrometer (Ocean Optics, Maya 2000) is used to verify the center wavelength of the generated light.

For this calibration, the objective lens was removed.
Before entering the optical system, the monochromator output was passed through an external diffuser (Thorlabs, ED1-C20-MD) used only to spatially homogenize the illumination.
The homogenized light then entered the system, passed through the system diffuser, the BRF, and the kaleidoscope, and was recorded by the camera.

We scan the monochromator at \SI{10}{\nano\metre} intervals from \SIrange{430}{670}{\nano\metre} and capture one image at each wavelength.
For each wavelength, we normalize each pixel in all replicated view-channels by the mean intensity of the BRF-free region in the central (zero-reflection) view-channel.
The normalized responses over wavelength are treated as pixel-wise measurements of the effective transmittance profiles corresponding to $B_v(\lambda)$ and are used to construct the per-pixel sensing matrix.

\subsection{Estimating Illumination--Sensor Response and Throughput Factors}
\label{ssec:supp_phi_g_estimation}

After calibrating the view-channel-wise BRF transmittance \(B_v(\lambda)\), we estimate the product of the illumination spectrum and the camera RGB sensitivities, \(\phi^c(\lambda)=l(\lambda)f^c(\lambda)\), together with wavelength-independent throughput factors \(g_v\), where \(v\) denotes the view-channel and \(c\in\{R,G,B\}\) denotes the sensor color channel.

For a white reference target, the observation of view-channel \(v\) and color channel \(c\) is modeled as

\begin{equation}
I_{v,\mathrm{w}}^{c}
=
g_v
\int_{\Lambda}
\phi^c(\lambda)\,B_v(\lambda)\,d\lambda
+\varepsilon_{v}^{c},
\label{eq:supp_white_obs}
\end{equation}
where \(g_v\) is a wavelength-independent throughput factor, and \(\varepsilon_{v}^{c}\) denotes the residual term accounting for measurement noise and model mismatch.
Discretizing \Cref{eq:supp_white_obs}, we obtain

\begin{equation}
I_{v,\mathrm{w}}^{c}
=
g_v\,\mathbf{b}_{v}^{\top}\phi^c
+\varepsilon_{v}^{c},
\label{eq:supp_white_disc}
\end{equation}
where \(\phi^c\in\mathbb{R}^{N_\lambda}\) is the discretized spectral response and
\((\mathbf{b}_{v})_m=B_v(\lambda_m)\Delta\lambda_m\).

With \(g_v\) fixed, we estimate \(\phi^c\) by
\begin{equation}
(\hat{\phi}^c,\hat{\alpha}_c)
=
\arg\min_{\phi^c,\alpha_c}
\sum_{v}
\bigl(
I_{v,\mathrm{w}}^c-g_v\,\mathbf{b}_{v}^{\top}\phi^c
\bigr)^2
+\beta_{\phi}\|\mathbf{D}\phi^c\|_2^2
+\gamma_{\mathrm{h}}\|\phi^c-\alpha_c\tilde{f}^c\|_2^2 ,
\label{eq:supp_phi_estimation}
\end{equation}
where \(\mathbf{D}\) is a difference matrix along wavelength, \(\tilde{f}^c\) is an independently measured relative camera-sensitivity shape prior, and \(\alpha_c\) absorbs the unknown scale of the prior. 
Here, the prior does not force \(\phi^c(\lambda)\) to coincide with the bare camera sensitivity \(f^c(\lambda)\). 
Rather, it encourages a similar spectral shape while allowing \(\phi^c(\lambda)\) to absorb additional wavelength-dependent effects introduced by the imaging optics, including the diffuser and the objective lens.
This is particularly necessary because the objective lens was removed during the calibration of \(B_v(\lambda)\), so its wavelength-dependent response is not included in \(B_v(\lambda)\) and is instead absorbed into \(\phi^c(\lambda)\).

With \(\phi^c\) fixed, we update the channel-wise throughput factors by regularized least squares and alternate these updates until convergence.
In practice, we use multiple observations for each view-channel and allow mild spatial variation in the gain correction with additional smoothness regularization.
Note that the absolute values of \(\beta_{\phi}\) and \(\gamma_{\mathrm{h}}\) are implementation-dependent.
In our implementation, the observation term is evaluated on the measured intensity scale, and its numerical scale is further affected by the wavelength discretization and the scaling of the calibrated transmittance and sensitivity terms.
Therefore, relatively large numerical values of \(\beta_{\phi}\) and \(\gamma_{\mathrm{h}}\) are required to balance the data term and the prior terms.
We used
\(\beta_{\phi}=8\times10^{5}\) and
\(\gamma_{\mathrm{h}}=8\times10^{5}\),
and ran the alternating optimization for 3 iterations.

The resulting \(\hat{\phi}^c(\lambda)\) and \(g_v\) are then used to construct the sensing matrix for subsequent reconstruction.

\subsection{Inter-Channel Registration}

We register the replicated view-channels using images of an asymmetric circle-grid calibration target captured through the system.
For each target image, we crop the triangular region corresponding to each view-channel using predefined triangle vertices obtained from a separate calibration image of the replicated views.
For channels with mirror-induced parity reversals, we apply the corresponding flip before feature detection.

We use the zero-reflection view-channel as the reference view and denote it by \(v=1\).
We then detect a \(4\times5\) asymmetric circle grid in each cropped view using OpenCV's circle-grid detector with a blob detector.
We aggregate the detected grid correspondences over multiple target images and estimate a homography from each view-channel to the reference view using random sample consensus (RANSAC).
The reference view is used as the common coordinate system, and its homography is therefore set to the identity.

The estimated inter-channel homographies are fixed after calibration and used to warp all view-channels into the common reference coordinate system before forming the observation vector for spectral reconstruction.

\section{Additional Comparison with Existing Snapshot HSI Systems}
\label{sec:supp_existing_systems}

Table~\ref{tab:supp_existing_systems} summarizes how the proposed system differs from representative snapshot HSI systems in terms of form factor, optical implementation, add-on compatibility, and reconstruction requirements.
The goal is not to exhaustively rank all systems, but to clarify the design trade-offs relevant to our target: a compact, low-cost, off-the-shelf camera add-on usable without supervised training.

\begin{table}[p]
\centering
\scriptsize
\setlength{\tabcolsep}{1.7pt}
\renewcommand{\arraystretch}{1.08}
\caption{
Qualitative comparison with representative snapshot HSI systems.
``Size / form factor'' lists reported dimensions when available, otherwise the reported prototype form factor; the Fujimoto et al.\ length is author-measured.
``Off-the-shelf,'' ``Camera add-on,'' and ``Train-free'' denote spectral-coding optics used as off-the-shelf components without custom fabrication, single-camera attachment without multi-camera or sensor-level modification, and reconstruction without supervised training, respectively.
}
\label{tab:supp_existing_systems}

\begin{tabularx}{\textwidth}{@{}
  >{\raggedright\arraybackslash}p{0.135\textwidth}
  >{\raggedright\arraybackslash}p{0.18\textwidth}
  >{\raggedright\arraybackslash}p{0.15\textwidth}
  >{\centering\arraybackslash}p{0.05\textwidth}
  >{\centering\arraybackslash}p{0.050\textwidth}
  >{\centering\arraybackslash}p{0.050\textwidth}
  >{\raggedright\arraybackslash}X
@{}}
\toprule
\multicolumn{1}{c}{Method}
& \multicolumn{1}{c}{\shortstack[c]{Size /\\ form factor}}
& \multicolumn{1}{c}{\shortstack[c]{Optical\\ strategy}}
& \multicolumn{1}{c}{\shortstack[c]{Off-the-\\shelf}}
& \multicolumn{1}{c}{\shortstack[c]{Camera\\ add-on}}& \multicolumn{1}{c}{\shortstack[c]{Train-\\free}}
& \multicolumn{1}{c}{\shortstack[c]{Trade-off\\ / note}} \\
\midrule

Baek et al.\ (2017)~\cite{Baek2017}
& DSLR + small front prism; dimensions not reported
& Prism dispersion over edges
& Yes
& Yes
& Yes
& Very compact front attachment, but relies on sparse edge cues and costly optimization. \\
\addlinespace[1.5pt]

Oh et al.\ (2016)~\cite{Oh2016_DIY}
& Three consumer cameras
& RGB cameras with distinct spectral sensitivities
& Yes
& No
& Yes
& Low-cost cameras, but requires multi-camera registration; dynamic scenes would need synchronized capture. \\
\addlinespace[1.5pt]

Takatani et al.\ (2017)~\cite{takatani2017one}
& Filter-attached faced mirrors: 300\,mm length, 10\,mm spacing
& Color-filter-attached faced reflectors
& Yes
& Yes
& Yes
& Very low-cost mirror attachment, but assumes flat scenes and repeated reflections reduce spatial resolution and SNR. \\
\addlinespace[1.5pt]

Manakov et al.\ (2013)~\cite{Manakov_2013_SnapHSI_Kaleido}
& Prototype: $\sim$1000\,mm incl.\ camera/main lens; multiplier: $36\times24\times300\,\mathrm{mm}$
& Kaleidoscope + filters + relay optics
& Partly
& Yes
& Yes
& Reconfigurable single-camera add-on, but relay optics increase length. \\
\addlinespace[1.5pt]

Zhao et al.\ (2019)~\cite{zhao2019hyperspectral}
& Relay prototype; compact sensor-mask version proposed
& Random printed mask + relay optics
& No
& Partly
& No
& Low-cost printed mask, but requires custom mask printing and CNN-based reconstruction. \\
\addlinespace[1.5pt]

Fujimoto et al.\ (2025)~\cite{Fujimoto2025}
& Relay prototype: $\sim$50\,cm total length
& Natural birefringent thin-section filter
& No
& Partly
& Yes
& No microfabrication, but reconstruction assumes locally uniform patches, reducing spatial resolution. \\
\addlinespace[1.5pt]

Gorman et al.\ (2010)~\cite{Gorman:10_BRFSnapHSI}
& Cage-mounted birefringent demultiplexer; system length not reported
& Cascaded birefringent interferometers + Wollaston demultiplexing
& No
& Partly
& Yes
& Direct spectral-image recording, but wider FoV/NA requires large, costly Wollaston-prism optics. \\
\addlinespace[1.5pt]

Kudenov et al.\ (2012)~\cite{Kudenov:12_BRFSnap}
& $15\times15\times10\,\mathrm{mm}^{3}$ common-path interferometer; benchtop prototype
& Nomarski-prism interferometer + lenslet arrays
& No
& No
& Yes
& Compact core optics, but specialized interferometric optics and multi-step calibration are required. \\
\addlinespace[1.5pt]

DOE/filter-array systems~\cite{Jeon2019,Yako2023,Bian2024}
& On-chip / sensor-level coding elements
& Custom DOE or filter-array coding
& No
& No
& Varies
& Compact, but relies on fabricated coding elements or sensor-level modification. \\

\addlinespace[2pt]
\midrule
\textbf{Ours}
& \textbf{21\,cm add-on; 40\,cm incl.\ camera}
& \textbf{Tapered kaleidoscope + directly attached BRF}
& \textbf{Yes}
& \textbf{Yes}
& \textbf{Yes}
& \textbf{No relay/cascade; directly attached off-the-shelf optics, but reduced reconstruction area from view replication.} \\
\bottomrule
\end{tabularx}
\end{table}

\section{Detailed Synthetic Experimental Setup}

We compare our method with two low-cost snapshot baselines, \textit{3Camera}~\cite{Oh2016_DIY} and \textit{Faced Reflectors}~\cite{takatani2017one}.
For all methods, the method-specific channel-wise spectral responses are used only in the forward model to generate synthetic observations, while spectral reconstruction is performed using the same non-negative smoothness-regularized least-squares formulation.
This protocol is intended to isolate the effect of the spectral responses used for measurement formation.

We generate synthetic observations from the 24 ColorChecker reflectance spectra on the same wavelength grid used in our system, namely \SIrange{430}{670}{\nano\metre} at \SI{10}{\nano\metre} intervals (25 wavelength samples).
To compare the methods under matched signal levels, we apply exposure matching so that, for each method, the maximum noise-free observation over all 24 patches and all observation entries is approximately 255\,digital numbers (DN).
After this exposure matching, we add i.i.d.\ Gaussian noise in the observation domain.

For each tested noise level, we sweep the smoothness weight \(\beta\) over \(0\) and logarithmically spaced values from \(10^{-2}\) to \(10^{6}\), and report the best peak signal-to-noise ratio (PSNR) over \(\beta\).
This evaluation is intended to compare the sensing designs under a common reconstruction model rather than to tune reconstruction settings separately for each method.
As in \Cref{ssec:supp_phi_g_estimation}, the absolute value of the smoothness weight is implementation-dependent.
In our implementation, the observation vector is represented on an exposure-matched DN scale, and its numerical scale is further affected by the wavelength discretization and by the scaling of the spectral response terms in the sensing matrix.
Therefore, relatively large numerical values of \(\beta\) are required to balance the regularization term.

We next describe the spectral responses used for each method in the forward model.
For the baseline methods, the spectral responses used in the forward model are digitized from spectral plots reported in the original papers.

\subsection{3Camera Baseline}

\begin{figure}[t]
    \centering
    \includegraphics[width=0.96\linewidth]{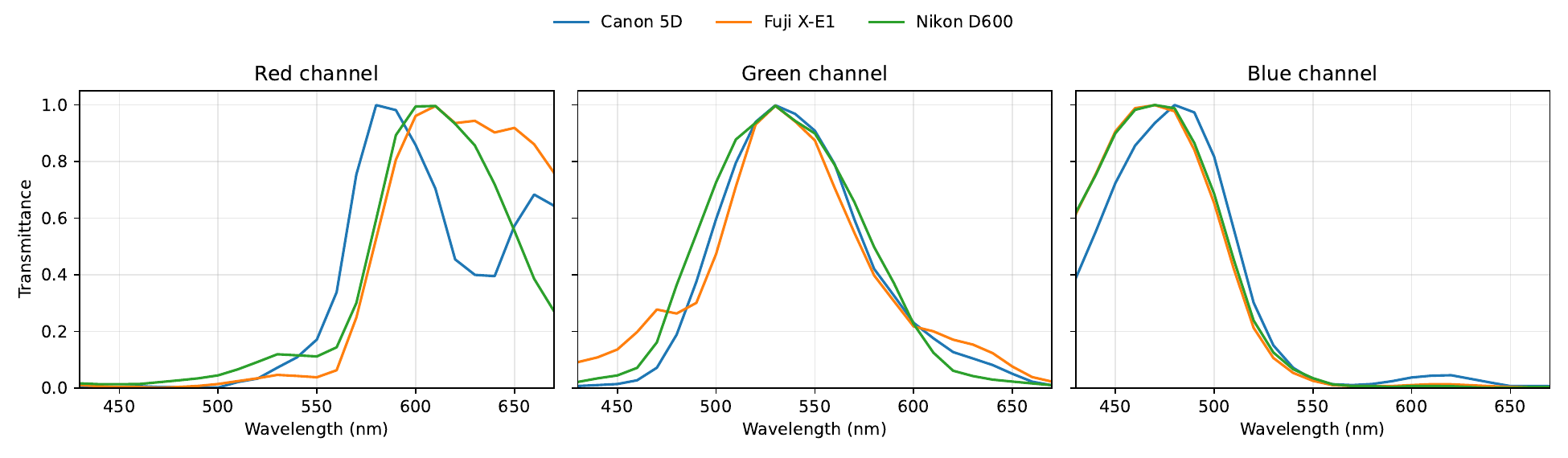}
    \caption{
    Camera spectral sensitivities used to model the \textit{3Camera} baseline~\cite{Oh2016_DIY}.
    }
    \label{fig:oh_filters}
\end{figure}

For the \textit{3Camera} baseline, we use the three-camera setting reported in~\cite{Oh2016_DIY}: Canon 5D, Fuji X-E1, and Nikon D600.
Each camera-color pair is treated as an independent measurement channel, yielding nine channels in total.
The corresponding spectral sensitivities are shown in \Cref{fig:oh_filters}.

Under the common linear model, the effective spectral response of camera \(q\) and color channel \(c\) is
\begin{equation}
T_{q,c}(\lambda)=f_q^c(\lambda),
\end{equation}
where \(q\in\{1,2,3\}\) indexes the cameras and \(c\in\{R,G,B\}\) indexes the sensor color channels.

\subsection{Faced Reflectors Baseline}

\begin{figure}[t]
    \centering
    \begin{subfigure}[t]{0.48\linewidth}
        \centering
        \includegraphics[width=\linewidth]{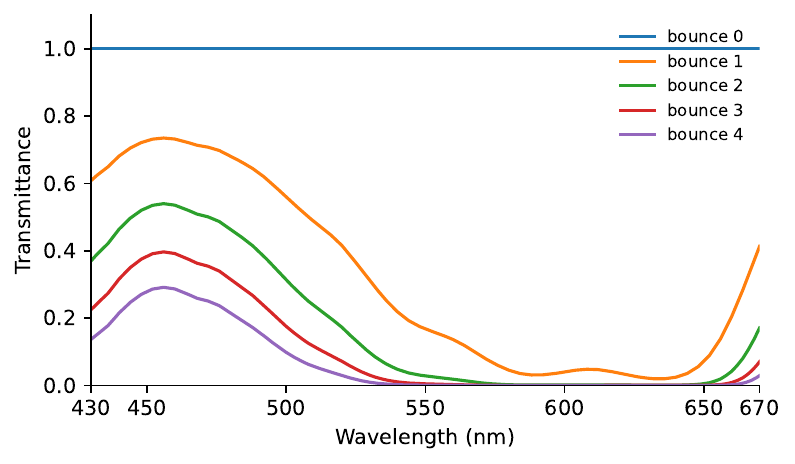}
        \caption{\#67 Light Sky Blue.}
        \label{fig:takatani_filters:a}
    \end{subfigure}
    \hfill
    \begin{subfigure}[t]{0.48\linewidth}
        \centering
        \includegraphics[width=\linewidth]{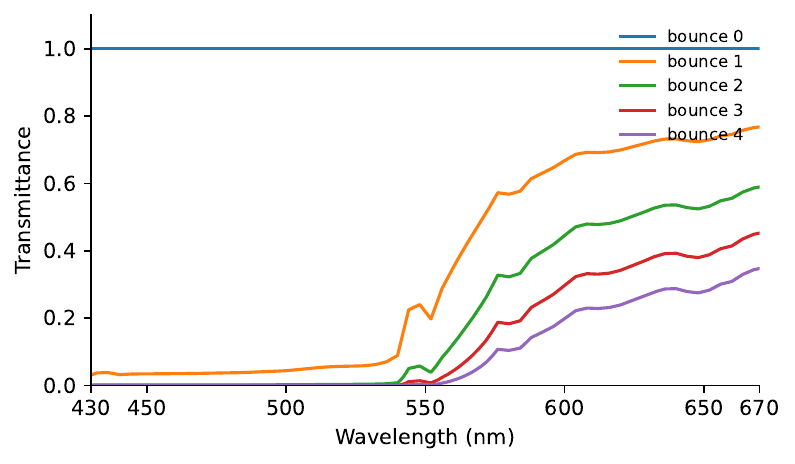}
        \caption{Thin clear orange filter.}
        \label{fig:takatani_filters:b}
    \end{subfigure}
    \caption{
    Bounce-dependent effective transmittances used to model the \textit{Faced Reflectors} baseline~\cite{takatani2017one}.
    }
    \label{fig:takatani_filters}
\end{figure}

For the \textit{Faced Reflectors} baseline, we follow the original bounce-dependent model in which the same filter is attached to the opposed reflectors.
If the base filter transmittance is \(\tau(\lambda)\), the \(i\)-th bounce is modeled by
\begin{equation}
T_{i,c}(\lambda)=f^c(\lambda)\,\tau(\lambda)^i,
\end{equation}
where \(f^c(\lambda)\) is the RGB sensor sensitivity and \(i\) is the bounce order.

In the synthetic evaluation, we use two filter settings from~\cite{takatani2017one}: \#67 Light Sky Blue, which was reported to provide favorable conditioning in the synthetic analysis, and the thin clear orange filter, which was used in their real-data prototype.
\Cref{fig:takatani_filters} shows the corresponding bounce-dependent transmittances used in the forward model.

\section{Experimental Validation of the OPD Distribution Using Real Data}

\begin{figure}[t]
    \centering
    \includegraphics[width=\linewidth]{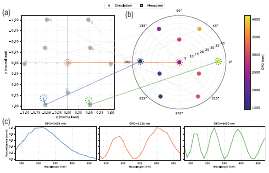}
    \caption{
    Validation of the OPD, $\Delta_{o-e}$, distribution from measured data.
    (a) Correspondence between measured points (squares) and simulated view-channels (circles) on the image plane.
    Each matched pair is connected by a line segment.
    (b) Estimated OPD values visualized on the $(\alpha,\sigma)$ polar parameterization of the view-channels, where color indicates the estimated $\Delta_{o-e}$.
    Matched channel pairs are again connected by line segments.
    (c) Examples of normalized transmittance spectra at the three dashed-circle locations highlighted in (a) and (b), corresponding to small, intermediate, and large OPD values.}
        \label{fig:OPD_Angle}
\end{figure}

\Cref{fig:OPD_Angle} compares the measured OPD estimates with the simulated view-channel geometry and shows representative transmittance spectra.
In the main text, the angular distribution of the OPD is analyzed primarily through simulation.
Here, we provide an experimental validation showing that the OPD distribution observed in the real system is qualitatively consistent with the simulated prediction.

For each measured view-channel, we use the calibrated spectral transmittance profile and estimate the OPD from its periodic modulation.
Because the birefringence-induced transmittance is periodic with respect to wavenumber rather than wavelength, we first convert the spectrum from wavelength $\lambda$ to wavenumber $k=1/\lambda$ and resample it on a uniformly spaced grid in $k$.
Under the transmittance model in the main text, the view-channel-wise spectral modulation can be written as
\begin{equation}
T(k)\propto \frac{1}{2}\left[1-\cos\!\left(2\pi\,\Delta_{o-e}\,k\right)\right],
\qquad
k=\frac{1}{\lambda},
\label{eq:supp_opd_fft_model}
\end{equation}
so that the dominant frequency in the wavenumber domain directly corresponds to the OPD $\Delta_{o-e}$.

We therefore estimate \(\Delta_{o-e}\) for each measured point by applying a 1D fast Fourier transform (FFT) to the uniformly resampled transmittance profile in the wavenumber domain and identifying the dominant spectral frequency component.
Independently, for each simulated view-channel, we compute its image-plane location and angular coordinates \((\alpha,\sigma)\) by ray tracing under the same tapered-kaleidoscope geometry as in the main text.
To compare simulation and measurement, we match each measured point to the nearest simulated view-channel on the image plane and transfer the estimated OPD values to the corresponding simulated angular coordinates \((\alpha,\sigma)\).

The measured OPD distribution is qualitatively consistent with the simulated prediction.
In particular, it exhibits approximate mirror symmetry and varies across view-channels.
These observations indicate that the replicated measurements provide angle-dependent spectral diversity in the real system.
Although this validation is qualitative and based on discrete channel samples, it remains consistent with the optical design assumption underlying the proposed system.

\section{Spatial Resolution Evaluation}
\label{sec:supp_spatial_resolution}

\begin{figure}[t]
\centering

\begin{subfigure}[t]{0.49\linewidth}
    \centering
    \includegraphics[width=\linewidth]{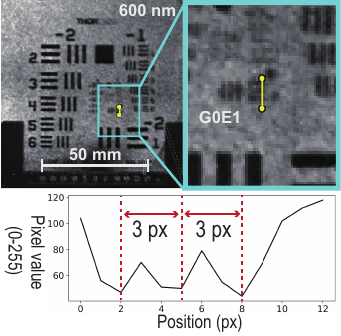}
    \caption{Reconstructed HSI at 600\,nm.}
    \label{fig:supp_usaf_resolution:hsi}
\end{subfigure}
\hfill
\begin{subfigure}[t]{0.49\linewidth}
    \centering
    \includegraphics[width=\linewidth]{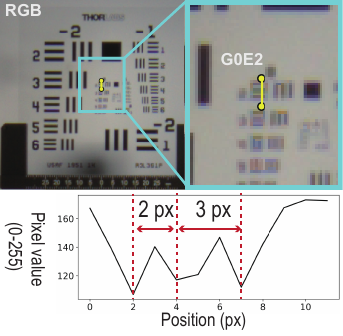}
    \caption{Without kaleidoscope/diffuser.}  
    \label{fig:supp_usaf_resolution:rgb_ref}
\end{subfigure}

\caption{
Spatial-resolution evaluation using a 1951 USAF target.
(a) Reconstructed HSI result at 600\,nm.
(b) RGB reference captured without the kaleidoscope/diffuser under the same objective lens, camera, and image-plane scale.
}
\label{fig:supp_usaf_resolution}
\end{figure}

This section provides additional details for the spatial-resolution evaluation summarized in the main paper.
We evaluate the effective spatial resolution of the prototype using a 1951 USAF resolution target (Thorlabs R3L3S1P).
\Cref{fig:supp_usaf_resolution} compares a representative reconstructed HSI band at 600\,nm with an RGB reference captured without the kaleidoscope/diffuser.

For the proposed-system measurement, the target was captured with the same prototype, calibration, inter-channel registration, reconstruction pipeline, and smoothness weight used for the real-scene reconstruction in the main paper.
The capture settings were \SI{0.25}{\second}, ISO 200, and F/8 under daylight-level artificial solar illumination.
We reconstructed 25 bands over \SIrange{430}{670}{\nano\metre} at \SI{10}{\nano\metre} intervals and evaluated the highest resolvable USAF group/element patterns.

For comparison, the RGB reference image was captured without the kaleidoscope/diffuser using the same objective lens and camera.
The object distance and image-plane scale were matched between the reconstructed HSI result and the RGB reference.
The RGB reference was captured at \SI{0.01}{\second}, ISO 200, and F/8.

The 1951 USAF target defines the spatial frequency of group \(G\) and element \(E\) as
\begin{equation}
\label{eq:supp_usaf_frequency}
r(G,E)
=
2^{G+(E-1)/6}
\quad
[\si{\linepair\per\milli\metre}].
\end{equation}

Most reconstructed HSI bands show resolvable structures at \SIrange{0.89}{1.00}{\linepair\per\milli\metre}, corresponding to Group \(-1\) Element 6 to Group 0 Element 1.
The representative 600\,nm band in \Cref{fig:supp_usaf_resolution:hsi} is consistent with this range.
The RGB reference without the kaleidoscope/diffuser in \Cref{fig:supp_usaf_resolution:rgb_ref} resolves \SI{1.12}{\linepair\per\milli\metre}, corresponding to Group 0 Element 2.
Thus, in this USAF test, the limiting resolution of the reconstructed HSI is within one to two USAF group/element steps of the RGB reference.
This indicates that the dominant spatial cost of the current prototype is the reduced valid reconstruction area from 10-view replication, rather than a large additional limiting-resolution loss caused by diffuser blur or inter-channel registration.

\noindent\textbf{Valid reconstruction region.}
We define the valid reconstruction region as pixels with at least three registered BRF-modulated view-channel observations.
Near the boundaries of the registered view-channel regions, not all ten view channels are available at every pixel.
For such pixels, unavailable observations are masked out, and the same pixel-wise reconstruction problem is solved using the corresponding available rows of the sensing matrix.
In the 10-view configuration, this valid region contains approximately \(3.24\times10^{5}\) pixels, corresponding to \SI{4.5}{\percent} of the bounding box enclosing the registered view-channel regions.



%
%
\bibliographystyle{splncs04}
\bibliography{main}